\documentclass[letterpaper]{article} 
\usepackage{aaai23}  
\usepackage{times}  
\usepackage{helvet}  
\usepackage{courier}  
\usepackage[hyphens]{url}  
\usepackage{graphicx} 
\usepackage{natbib}  
\usepackage{caption} 
\usepackage{algorithm}
\usepackage[noend]{algorithmic}
\usepackage{threeparttable}
\usepackage{newfloat}
\usepackage{listings}
\DeclareCaptionStyle{ruled}{labelfont=normalfont,labelsep=colon,strut=off} 
\floatstyle{ruled}
\newfloat{listing}{tb}{lst}{}
\floatname{listing}{Listing}
\usepackage{url}
\usepackage{graphicx}
\usepackage{bm}
\usepackage{color}
\usepackage{multirow}
\usepackage{amssymb}
\usepackage{amsmath}

\begin{document}

\title{Compiler Provenance Recovery for Multi-CPU Architectures Using a Centrifuge Mechanism}
\author{Yuhei Otsubo\textsuperscript{\rm 123}, Akira Otsuka\textsuperscript{\rm 3}, Mamoru Mimura\textsuperscript{\rm 4}}
\affiliations{
\textsuperscript{\rm 1} National Police Academy, 3-12-1 Asahi-cho, Fuchu, Tokyo, Japan\\
dgs157101@iisec.ac.jp\\
\textsuperscript{\rm 2} National Police Agency, 2-1-2 Kasumigaseki, Chiyoda, Tokyo, Japan\\
\textsuperscript{\rm 3} Institute of Information Security, 2-14-1 Tsuruya-cho, Kanagawa, Yokohama, Kanagawa, Japan\\
\textsuperscript{\rm 4} National Defense Academy, 1-10-20 Hashirimizu, Yokosuka, Kanagawa, Japan\\
}
\maketitle

\begin{abstract}
\begin{quote}
Bit-stream recognition (BSR) has many applications, such as forensic investigations, detection of copyright infringement, and malware analysis.
We propose the first BSR that takes a bare input bit-stream and outputs a class label without any preprocessing. To achieve our goal, we propose a centrifuge mechanism, where the upstream layers (sub-net) capture global features and tell the downstream layers (main-net) to switch the focus, even if a part of the input bit-stream has the same value.
We applied the centrifuge mechanism to compiler provenance recovery, a type of BSR, and achieved excellent classification.
Additionally, downstream transfer learning (DTL), one of the learning methods we propose for the centrifuge mechanism, pre-trains the main-net using the sub-net's ground truth instead of the sub-net's output.
We found that sub-predictions made by DTL tend to be highly accurate when the sub-label classification contributes to the essence of the main prediction.

\end{quote}
\end{abstract}

\section{Introduction}
Bit-stream recognition (BSR) is a classification task that takes a bit-stream as input and outputs a class label.
In the field of cybersecurity, which is the practice of protecting systems, networks, and programs from digital attacks, various studies have used methods involving BSR models, including
the classification of malicious programs~\cite{nataraj2011malware}, author identification of programs~\cite{Rosenblum2011WhoWT}, discovering vulnerabilities in programs~\cite{padmanabhuni2015buffer}, function recognition~\cite{rosenblum2007machine}, and recovering corrupted files~\cite{Heo2019}.
Bit-streams sometimes appear to be very complicated, as they are mostly produced by artificial generative models such as program compilers and audio codecs.
Because the bit-stream structure varies greatly depending on the artificial generative model and the location or context from which the bit-stream was taken, data engineering (DE) is among the most significant steps for achieving classification accuracy.

To our knowledge, no BSR model that uses raw bit-streams as training data and has high classification accuracy has yet been developed.
Compiler provenance recovery (CPR), one of the applications of BSR, refers to the task of identifying the environment in which given program binaries were created. When CPR is applied to malware analysis, it can provide important information for determining the author of the malware.
CPR is mainly a classification for machine language instruction sequences. However, because the model cannot be trained efficiently with raw bit-streams, various methods for creating feature vectors have been proposed, including replacing disassembled machine language instructions with several categories of symbols to see the flow of program code~\cite{rahimian2015bincomp} and creating feature vectors on a per-function basis~\cite{rosenblum2010extracting,rosenblum2011recovering}.
Otsubo et al. proposed o-glassesX~\cite{otsuboglassesx}, which can be trained with data that are relatively close to raw bit-streams, and then classification can be performed with high accuracy by dividing the bit-stream into instruction units in pre-processing.

Most existing BSR models are domain dependent and are designed to improve classification accuracy by narrowing down the input bit-stream target domain and specializing the DE for a particular purpose.
Similarly in CPR, the key to highly accurate classification is to limit the corresponding CPU architecture and DE according to its type.
No generic BSR exists that can apply existing models to other BSR problems without redesigning the DE to achieve high classification performance.
Therefore, in this study, we aim to propose a CPR method without DE that can classify bit-streams with high accuracy.

We propose a centrifuge mechanism in which the upstream sub-net transitions the input to a space corresponding to sub-labels.
Our method is the first to enable highly accurate BSR without DE.
The key idea of the centrifuge mechanism is to utilize a sub-net predictor that captures global features to automate the process of narrowing down the target domain, which has been done manually in the past.
The model incorporates a sub-net predictor that captures global features and recognizes the difference, even if the same bit-blocks appear in different bit-streams from various sources.
The centrifuge allows the downstream main-net to focus on more difficult classifications.
For the centrifuge mechanism, we checked the accuracy and characteristics of the main-net and sub-net predictions using various learning methods.
One of them pre-trains the main-net using the sub-label's ground truth instead of the sub-net's output.
This method was able to give the sub-net the role of sub-label prediction without using a loss function for sub-label classification.
Additionally, we found that sub-predictions tend to be highly accurate when the sub-label classification contributes to the essence of the main prediction.

The contributions of this paper are as follows.
\begin{itemize}
    \item To our knowledge, our proposed model  is the first CPR model able to simultaneously support multiple CPU architectures, and it has achieved cutting edge performance for CPR in terms of classification performance.

    \item We demonstrated that a single loss function can be used to generate a sub-net in the model that allow subclass classification independent of the main class.

    \item We proposed a new learning method that improves the interpretability of the output by explicitly assigning roles in the model.

\end{itemize}

\section{Related work}
\label{sec:related}
To our knowledge, no BSR model that uses raw bit-streams as training data and has high classification accuracy has yet been developed.
In this section, we describe existing work on CPR as an example of BSR problems.
In this paper, we experimentally demonstrate our model's applicability to CPR.

Several compiler identification tools are already available (i.e., IDA Pro\footnote{\url{https://www.hex-rays.com/products/ida/}}, PEiD\footnote{\url{https://www.aldeid.com/wiki/PEiD}}, and RD\footnote{\url{http://www.rdgsoft.net/}}). These tools are roughly signature-based and typically rely on metadata or other details in the program header. Their exact matching algorithm may fail if even a slight difference between signatures is present, or if the header information has been stripped or is otherwise unavailable.

Rosenblum~\emph{et al.} used a conditional random field (CRF~\cite{mccallum2002efficiently}) and set up a classifier that took a bit-stream as input to identify one of the three compiler families with a 0.924 accuracy rate~\cite{rosenblum2010extracting}.
They disassembled the bit-stream with an IA-32 architecture and found a typical matching instruction pattern called ``idioms'' that predicted the compiler families. Their early result is only for the relatively easy compiler family identification with three classes.
Rosenblum~\emph{et al.} have improved their method and proposed a tool named ORIGIN~\cite{rosenblum2011recovering}.
ORIGIN's linear support vector machine~\cite{cortes1995support} takes the feature of an independent ``function'' as input and predicts both the optimization level and the version in addition to the compiler family, but it has difficulty identifying the compiler versions. ORIGIN's CRF takes the same features of multiple adjacent functions as input for the prediction and performs with a higher accuracy of 0.9 and above, despite the number of classes having increased from 3 to 18. However, the size of the input data required for predicting compiler provenance is larger.

Rahimian~\emph{et al.} developed BinComp~\cite{rahimian2015bincomp}, an approach in which the syntax, structure, and semantics of disassembled functions are analyzed to extract the compiler provenance.
In their experiments, BinComp had an identification accuracy of 0.801 in 8-class classification.

o-glasses~\cite{otsubo2018glasses} and o-glassesX~\cite{otsuboglassesx} have succeeded in identifying compilers with very high accuracy by dividing the raw bit-stream into x86/x86-64 machine language instruction units and combining a convolutional neural network (CNN) and an attention module.
These two methods can identify the compiler family and the optimization level and version with high accuracy merely by analyzing instruction sequences without regard to functions.

All of these existing studies require pre-processing, such as disassembly of machine language instructions and structural analysis of functions, to achieve their high precision classifications.
We show that our method achieves high accuracy classification without pre-processing.

\section{Centrifuge mechanism}
\label{sec:centrifuge}
The structure of bit-streams differs greatly depending on the type of file.
Even if we consider only the case of machine language instruction strings, a general linear layer cannot transform the input bit-stream into the instruction vector sequence since bit-blocks with the same value have different meanings on different CPU architectures.
We therefore propose a centrifuge mechanism that transforms local features like bit-blocks into a feature vector by considering global features.

\begin{figure}[!tb]
\begin{center}
\includegraphics[width=\columnwidth]{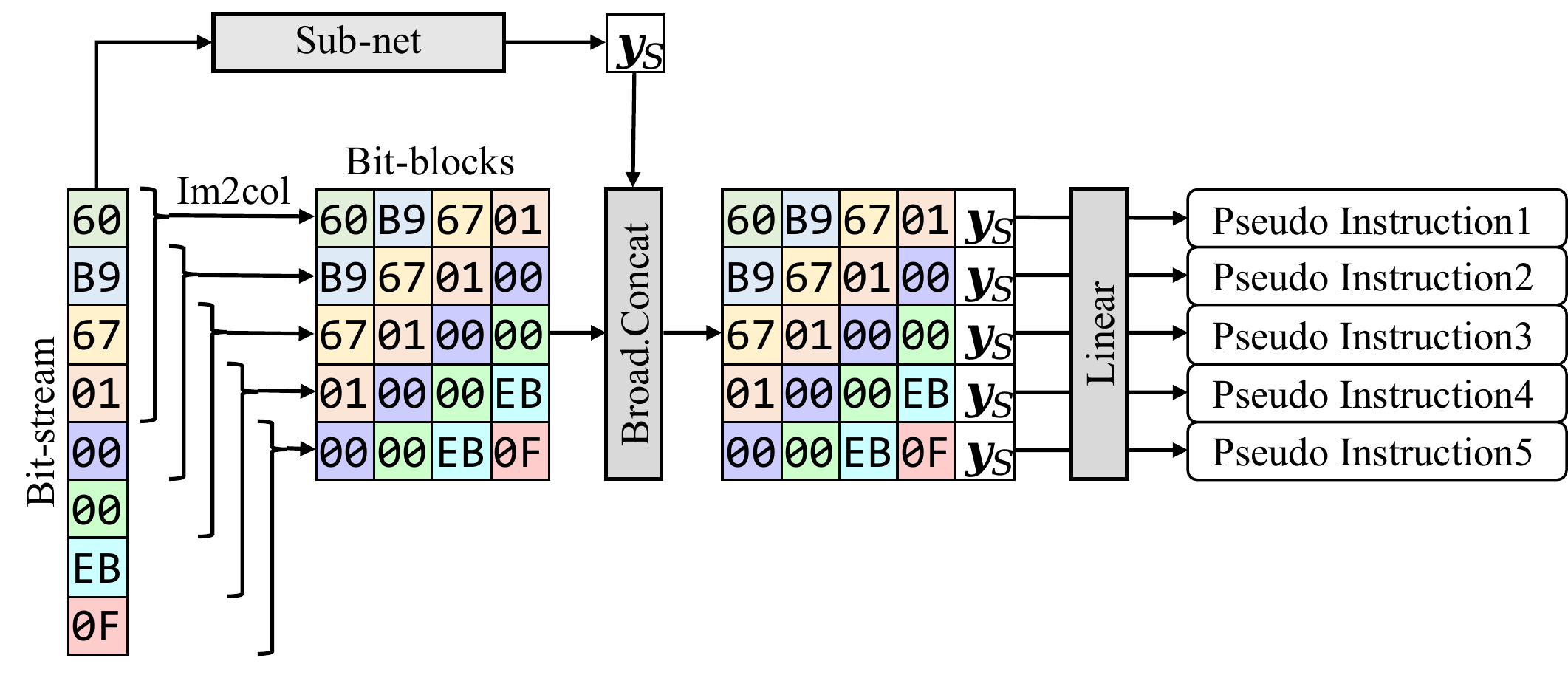}
\caption{Instruction embedding using a sub-net and linear.}
\label{fig:embedding}
\end{center}
\end{figure}

Figure~\ref{fig:embedding} shows an example of the behavior of the centrifuge mechanism on the learning program code of various CPU architectures.
The centrifuge mechanism consists of a sub-net that learns global features and a broadcast concatenator.
The broadcast concatenator concatenates an input of the sub-net and the $\mathbf{y}_\mathrm{S}$(the sub-net's output.)
This concatenation changes operations that use local features of the main net to those that consider global features.

Even bit-blocks with the same value can have different meanings if the CPU architecture is different.
Assuming that the sub-net learns the global features such as the CPU architecture of the input bit-stream, we expect the output of the linear layer after sub-net embedding to produce a feature vector sequence similar to the instruction vector sequence.

\begin{figure}[!tb]
  \begin{minipage}[b]{.57\columnwidth}
    \begin{center}
    \includegraphics[width=0.9\columnwidth]{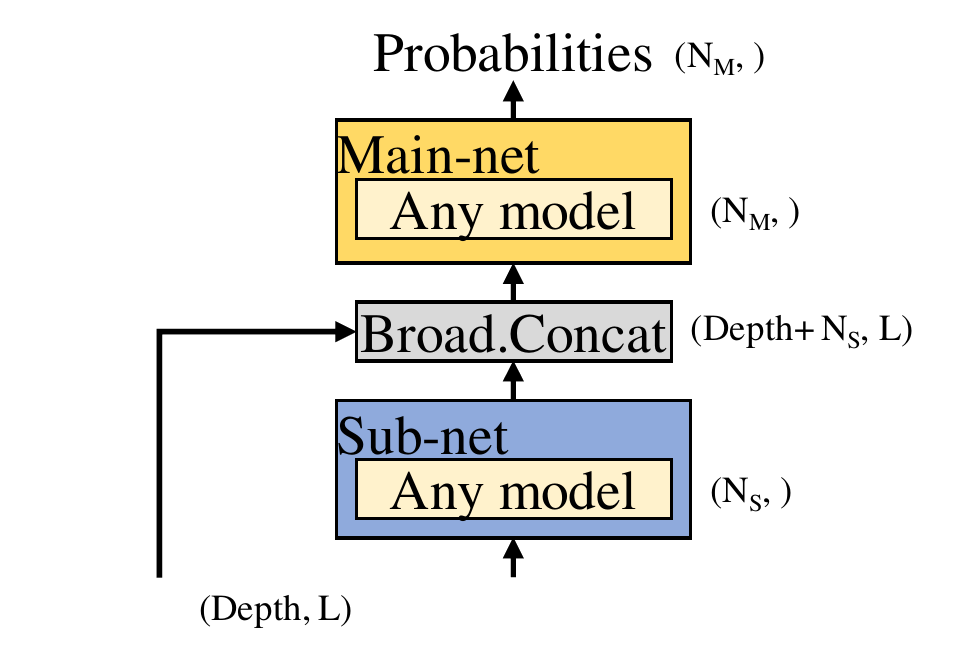}
    \caption{Outline of the centrifuge mechanism.}
    \label{fig:simple-model}
    \end{center}
  \end{minipage}
  \hspace{0.02\columnwidth}
  \begin{minipage}[b]{.37\columnwidth}
    \begin{center}
    \includegraphics[width=0.7\columnwidth]{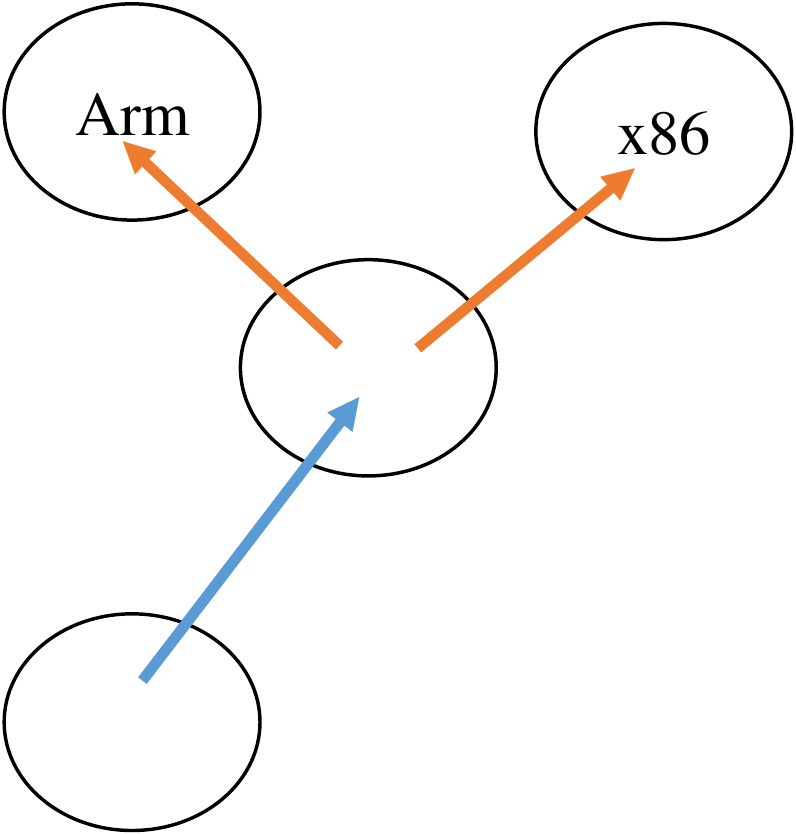}
    \caption{Transition of the input $\mathbf{x}$ by the first linear layer of the main-net.}
    \label{fig:x_dash}
        
    \end{center}
  \end{minipage}
\end{figure}

A basic form of the centrifuge mechanism is shown in Figure~\ref{fig:simple-model}.
Any model can be set up as a main-net and sub-net.
When the input of the sub-net is $\mathbf{x}_2$ and the sub-net model is $\mathrm{Sub}$ with parameter $\mathbf{w}_\mathrm{S}$, the output $\mathbf{y}_\mathrm{S}$ is expressed by
\begin{equation}
    \mathbf{y}_\mathrm{S} =\mathrm{Sub}(\mathbf{x}_2;\mathbf{w}_\mathrm{S}).
\end{equation}
When an operator $\oplus$ represents a broadcast concatenator operation, the input of the main-net is $\mathbf{x}_1\oplus\mathbf{y}_\mathrm{S}$.
Then, the output $\mathbf{y}_\mathrm{M}$ is expressed by
\begin{equation}
    \mathbf{y}_\mathrm{M} = \mathrm{Main}(\mathbf{x}_1\oplus\mathbf{y}_\mathrm{S};\mathbf{w}_\mathrm{M}),
\end{equation}
where the main-net model is $\mathrm{Main}$ with parameter $\mathbf{w}_\mathrm{M}$.

We assume that the first layer of the main-net is a linear layer.
When the parameter of the linear layer is $\{\mathbf{w}_1,\mathbf{w}_2\}$, the output of the linear layer $\mathbf{x}'$ is given by
\begin{equation}
\label{eq:x_dash}
    \mathbf{x}'=\mathrm{Linear}(\mathbf{x}_1\oplus\mathbf{y}_\mathrm{S})
    =\mathbf{x}_1\mathbf{w}_1+\mathbf{y}_\mathrm{S}\mathbf{w}_2.
\end{equation}
Then, assuming that the sub-net has acquired global features like the CPU architecture, $\mathbf{x}'$ is expected to be a transition of input x to the space for each global feature, a process shown in Figure~\ref{fig:x_dash} that resembles the action of a centrifuge.

$\mathbf{x}_2$ can take any value regardless of $\mathbf{x}_1$.
In this paper, we refer to the centrifuge when $\mathbf{x}_2$ and $\mathbf{x}_1$ are equal as the self-centrifuge and when $\mathbf{x}_2$ and $\mathbf{x}_1$ are different as the source-target centrifuge.

For the sub-net to explicitly acquire global features such as the CPU architecture, either a loss function or a learning method needs to be devised.
Next, we provide an overview of transfer learning, fine-tuning, and an objective function combining two loss functions.

\subsection{Transfer learning}
Transfer learning~\cite{pan2009survey} is a learning method that focuses on storing knowledge gained while solving a problem and applying it to a different but related problem. 

\subsubsection{Upstream transfer learning: pre-training sub-net}
Empirically, models such as AlexNet that are trained on vast amounts of labeled data such as Imagenet learn generic features in the layer close to the input (upstream).
These features are also effective in other tasks, so transferring the weights of this learned model can reduce training time and create a highly accurate model when there are little labeled data in the destination environment.
A common approach to transfer learning is to work with a trained model as a feature extractor.
We name this approach upstream transfer learning (UTL) to distinguish it from the downstream-focused approach described in the next section, 

To apply UTL to our model, we pre-train the upstream sub-net to work them as feature extractors.
When the ground truth for the subclass is $\mathbf{t}_\mathrm{S}$ and the loss function for the subclass is $\mathcal{L}_\mathrm{S}$, the pre-learning of the sub-net is given by
\begin{equation}
\label{eq:sub-train}
    \mathbf{w}_\mathrm{S}:=\mathbf{w}_\mathrm{S}-\alpha \left(\frac{\partial\mathcal{L}_\mathrm{S}(\mathbf{y}_\mathrm{S},\mathbf{t}_\mathrm{S})}{\partial\mathbf{w}_\mathrm{S}}+\lambda\mathbf{w}_\mathrm{S}\right).
\end{equation}

After completing the pre-training, the parameters of the main-net are updated using
\begin{equation}
\label{eq:main-train}
    \mathbf{w}_\mathrm{M}:=\mathbf{w}_\mathrm{M}-\alpha\left(\frac{\partial\mathcal{L}_\mathrm{M}(\mathbf{y}_\mathrm{M},\mathbf{t}_\mathrm{M})}{\partial\mathbf{w}_\mathrm{M}}+\lambda\mathbf{w}_\mathrm{M}\right),
\end{equation}
where $\mathbf{t}_\mathrm{M}$ is the ground truth for the main class and $\mathcal{L}_\mathrm{M}$ is the loss function for the main class.

\subsubsection{Downstream transfer learning: pre-training main-net}
In this section, we describe transfer learning focusing on downstream (downstream transfer learning [DTL]), in contrast to the upstream-focused approach in the previous section.
When the upstream weights of the network are fixed, the upstream works as a feature extractor.
In contrast, if we fix the weights downstream, near the exit layer of the network, then the downstream can be viewed as a loss function in complex formulas.

To apply DTL to the centrifuge mechanism, we pre-train the downstream main-net to work them as a loss function for the sub-net.
The DTL algorithm is described in Algorithm~\ref{alg:algorithm} and is given in detail below.
\begin{algorithm}[tb]
\caption{DTL algorithm}
\label{alg:algorithm}
\begin{algorithmic}[1] 
\renewcommand{\algorithmicrequire}{\textbf{Input:}}
\renewcommand{\algorithmicensure}{\textbf{Output:}}
\REQUIRE Loss function $\mathcal{L}_\mathrm{M}$, main-net predictor $\mathrm{Main}$, sub-net predictor $\mathrm{Sub}$, training dataset $S:=\bigcup^{n}_{i=1}\{(\mathbf{x}_{1i},\mathbf{x}_{2i},\mathbf{t}_{\mathrm{S}i},\mathbf{t}_{\mathrm{M}i})\}$, mini-batch size $b$, weight decay coefficient $\lambda$, scheduled learning rate $\alpha$, initial weight $\mathbf{w}_0$.
\ENSURE Trained weight $\mathbf{w}=\{\mathbf{w}_\mathrm{S},\mathbf{w}_\mathrm{M}\}$
\STATE Initialize weight $\mathbf{w}=\mathbf{w}_0$.
\\\COMMENT {Main-net training}
\WHILE{not converged}
\STATE Sample a mini-batch $B$ of size $b$ from $S$.\\
\STATE $\mathbf{y}_\mathrm{M}:=\mathrm{Main}(\mathbf{x}_1\oplus\mathbf{t}_\mathrm{S};\mathbf{w}_\mathrm{M})$
\STATE $\mathcal{L}_\mathrm{all}:=\mathcal{L}_\mathrm{M}(\mathbf{y}_\mathrm{M},\mathbf{t}_\mathrm{M})$
\STATE $\mathbf{w}_\mathrm{M}:=\mathbf{w}_\mathrm{M}-\alpha(\frac{\partial\mathcal{L}_\mathrm{all}}{\partial\mathbf{w}_\mathrm{M}}+\lambda\mathbf{w}_\mathrm{M})$
\ENDWHILE
\COMMENT {Sub-net training}
\WHILE{not converged}
\STATE Sample a mini-batch $B$ of size $b$ from $S$.\\
\STATE $\mathbf{y}_\mathrm{S}:=\mathrm{Sub}(\mathbf{x}_2;\mathbf{w}_\mathrm{S})$
\STATE $\mathbf{y}_\mathrm{M}:=\mathrm{Main}(\mathbf{x}_1\oplus\mathbf{y}_\mathrm{S};\mathbf{w}_\mathrm{M})$
\STATE $\mathbf{w}_\mathrm{S}:=\mathbf{w}_\mathrm{S}-\alpha(\frac{\partial\mathcal{L}_\mathrm{all}}{\partial\mathbf{w}_\mathrm{S}}+\lambda\mathbf{w}_\mathrm{S})$
\ENDWHILE
\STATE \textbf{return} $\mathbf{w}$
\end{algorithmic}
\end{algorithm}

First, fix the value of $\mathbf{y}_\mathrm{S}$ to $\mathbf{t}_\mathrm{S}$ for main-net pre-training,
\begin{equation}
    \mathbf{y}_\mathrm{M} = \mathrm{Main}(\mathbf{x}_1\oplus\mathbf{t}_\mathrm{S};\mathbf{w}_\mathrm{M}).
\end{equation}
The pre-learning of the main-net is indicated by Equation~(\ref{eq:main-train}).
After the pre-training of the main-net is completed, its parameters are fixed, and the sub-net is trained with the loss function of the main-net, as shown by
\begin{equation}
\label{eq:dtl}
    \mathbf{w}_\mathrm{S}:=\mathbf{w}_\mathrm{S}-\alpha\left(\frac{\partial\mathcal{L}_\mathrm{M}(\mathbf{y}_\mathrm{M},\mathbf{t}_\mathrm{M})}{\partial\mathbf{w}_\mathrm{S}}+\lambda\mathbf{w}_\mathrm{S}\right).
\end{equation}

When we put $\mathcal{L'}_\mathrm{S}(\mathbf{y}_\mathrm{S})\equiv \mathcal{L}_\mathrm{M}(\mathrm{Main}(\mathbf{x}\oplus\mathbf{y}_\mathrm{S};\mathbf{w}_\mathrm{M}),\mathbf{t}_\mathrm{M})$, Equation~\ref{eq:dtl} can be expressed as
\begin{equation}
    \mathbf{w}_\mathrm{S}:=\mathbf{w}_\mathrm{S}-\alpha\left(\frac{\partial\mathcal{L'}_\mathrm{S}(\mathbf{y}_\mathrm{S})}{\partial\mathbf{w}_\mathrm{S}}+\lambda\mathbf{w}_\mathrm{S}\right).
\end{equation}

Because the weights of the main-net are fixed, the shape of the $\mathcal{L'}_\mathrm{S}$ function is constant during learning.
In other words, $\mathcal{L'}_\mathrm{S}$ serves as a loss function from the standpoint of the sub-net.

\subsection{Fine-tuning}
Fine-tuning~\cite{girshick2014rich} modifies the weights of an existing model to train a new task.
The output layer is usually extended with randomly initialized weights for the new task.
A small learning rate is then used to tune all parameters from their original values to minimize the loss of the new task.
Part of the network is sometimes frozen to prevent overfitting. 

Fine-tuning, like transfer learning, also classifies the learned model into upstream fine-tuning and downstream fine-tuning depending on the location to which it is applied.

\subsubsection{Upstream fine-tuning: pre-training sub-net}
The same as for UTL, the sub-net is pre-trained according to Equation~(\ref{eq:sub-train}).
After completing the pre-training, update the model's parameters using the loss function for the main-net:

\begin{equation}
\label{eq:all-train}
    \mathbf{w}:=\mathbf{w}-\alpha\left(\frac{\partial\mathcal{L}_\mathrm{M}(\mathbf{y}_\mathrm{M},\mathbf{t}_\mathrm{M})}{\partial\mathbf{w}}+\lambda\mathbf{w}\right),
\end{equation}
where the parameters of the model are $\mathbf{w}=\{\mathbf{w}_\mathrm{S},\mathbf{w}_\mathrm{M}\}$.

\subsubsection{Downstream fine-tuning: pre-training sub-net}
The same as for DTL, the main-net is pre-trained according to Equation~(\ref{eq:main-train}).
After completing the pre-training, update the parameters of the model with Equation~(\ref{eq:all-train}).

\subsection{2LF: Learning with two loss functions}
We next describe a method for incorporating both the loss function for the main-net and the loss function for the sub-net into the overall model loss function.
There are various methods of incorporation, but the simplest is given by
\begin{equation}
    \mathcal{L}_\mathrm{all}=\mathcal{L}_\mathrm{M}(\mathbf{y}_\mathrm{M},\mathbf{t}_\mathrm{M})+\beta \mathcal{L}_\mathrm{S}(\mathbf{y}_\mathrm{S},\mathbf{t}_\mathrm{S}),
\end{equation}
where $\beta$ is a hyperparameter whose value determines whether the classification accuracy of the main-predictor or the sub-predictor is preferred.
However, the appropriate value of $\beta$ depends on other training conditions, such as the training data set.

\section{Experimental results}
Given that CPR is a motivation for our research, here we applied the centrifuge mechanism to CPR and observed its effectiveness in experiments.

There are three main objectives of the experiment:

\begin{itemize}
    \item To confirm the performance and characteristics of each of the five learning methods considered for the centrifuge mechanism (Ex.~1 to 3.).
    \item To confirm the performance and characteristics of the centrifuge mechanism when the number of sub-nets is increased from one to two (Ex.~4.).
    \item To compare the performance of existing CPR methods with DE with that of our experimental model without DE (Ex.~5.).
\end{itemize}

\subsection{Experimental model}

\begin{figure}[!tb]
\begin{center}
\includegraphics[width=\columnwidth]{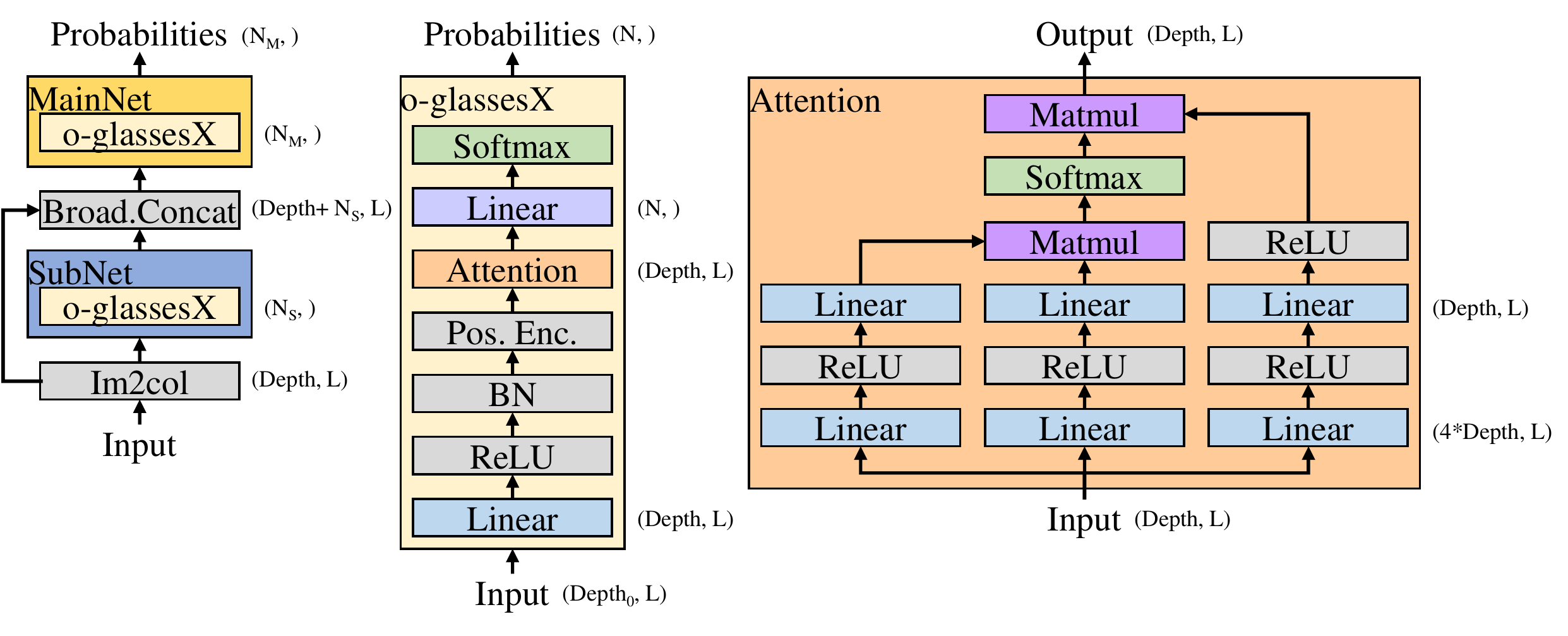}
\caption{Outline of our experimental model.}
\label{fig:model}
\end{center}
\end{figure}

In this section, we describe the model used in our experiments.
An overview of the experimental model is shown in Figure~\ref{fig:model}.
We selected self-centrifuge as the centrifuge mechanism and o-glassesX~\cite{otsuboglassesx} as the model used for the main-net and sub-net.
The reasons for selecting each are described below.

\subsubsection{o-glassesX}
o-glassesX, described in Section~\ref{sec:related}, has so far shown the best performance in the area of CPR.
However, o-glassesX assumes a single CPU architecture as the classification target since it performs preprocessing according to that architecture.
Specifically, preprocessing converts the input bit-stream into a fixed-length machine language instruction sequence, and a tool called a disassembler for each CPU architecture is essential for this conversion.
In our experiments, we replaced this CPU architecture-specific preprocessing with the centrifuge mechanism and used raw bit-streams as input.

\subsubsection{Self-centrifuge}
Some recent malware attacks do not save an executable file but instead deploy the executable code in a temporary memory area, making it difficult to trace the attacker.
In such cases, only a small portion of the executable file is available, and it is necessary to investigate the traces of the attacker from the file fragments.
We decided to select the self-centrifuge, which inputs the same program code to both the target and source to allow compiler identification even from file fragments.

On the other hand, if the entire executable file is available, the source-target centrifuge, where the file header area is input to the target ($\mathbf{x}_2$) and the program code to the source ($\mathbf{x}_2$), is considered appropriate.
The header area of the executable file contains various metadata, including information about the compiler and CPU architecture.
The compiler information is not necessary for executing the executable.
An attacker can easily tamper with the compiler information, while the CPU architecture information is essential for executing the executable and is difficult to tamper with.

\begin{table}[!tb]
\caption{Overview of our dataset.} 
\label{tab:dataset}
\begin{center}
  \includegraphics[width=\columnwidth,clip]{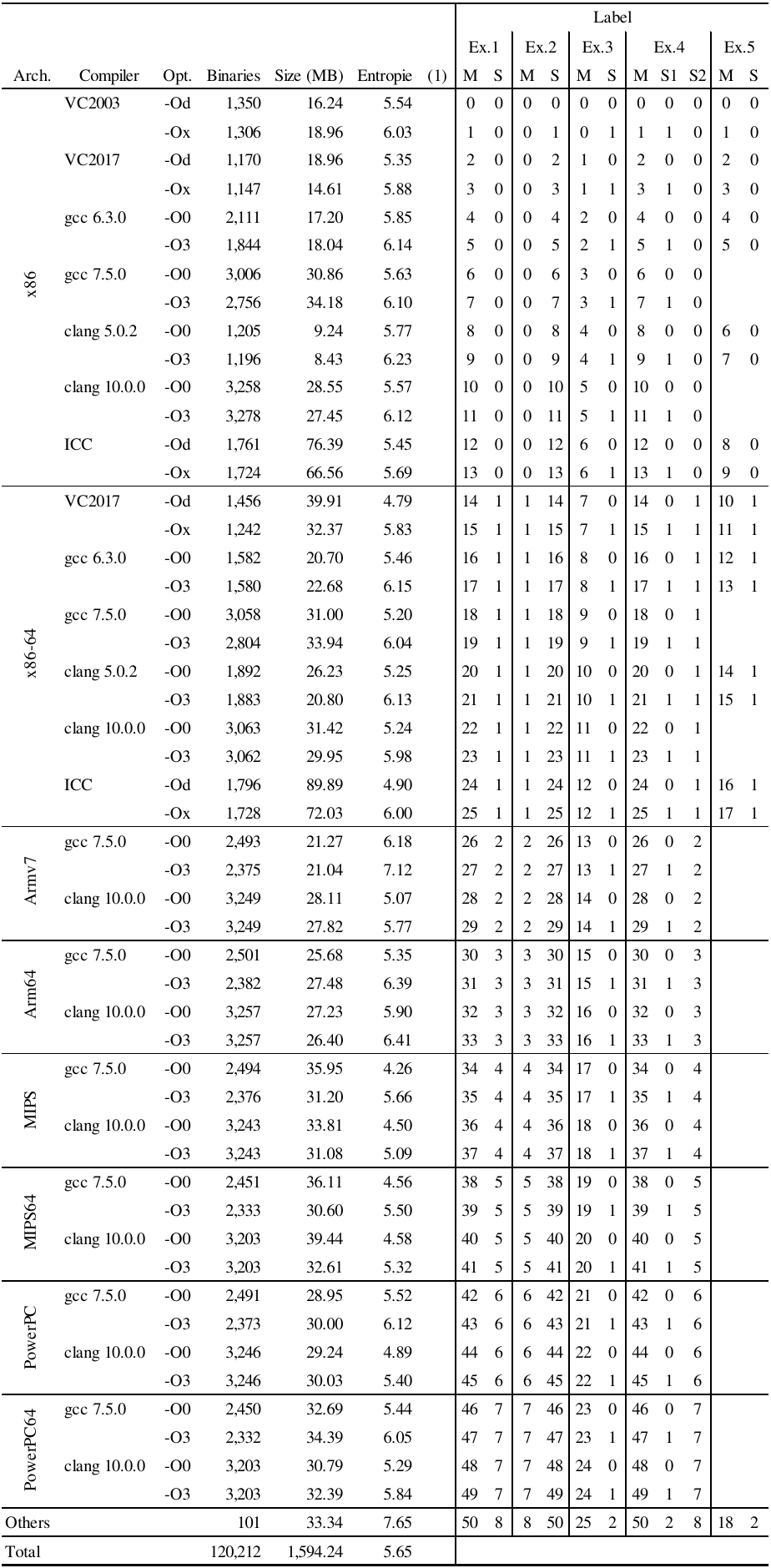}
    \begin{tablenotes}
     \footnotesize
     \item[*](1) is the same subset as a subset of the dataset of o-glassesX~\cite{otsuboglassesx}.
    \end{tablenotes}
\end{center}
\end{table}

\subsection{Dataset}
\label{sec:dataset}
To our knowledge, no dataset for CPR supports multiple CPU architectures.
Therefore, we created the dataset ourselves based on the dataset used in the o-glassesX~\cite{otsuboglassesx} experiments for enabling performance comparisons with existing methods that support only x86/x86-64.

The o-glassesX dataset is a compiler identification dataset and consists mainly of binaries for x86/x86-64 architectures.
First, we prepared the same C/C++ source code files from the o-glassesX dataset.
We then compiled these source code files with various compiler settings for multiple CPU architectures. Next, we extracted various bits of CPU architecture native code from the generated object files using {\tt elf\_coff2bin.py}, which was published with o-glassesX.
The size of our dataset was about 1.6~GB, and
Table~\ref{tab:dataset} shows an outline of the dataset.
The dataset consists of the native code of each architecture except ``Others.''
``Others'' is non-native code data created from various document files (.rtf, .doc, .docx, and .pdf files).
``Label'' in Table~\ref{tab:dataset} indicates the labels used in the five experiments described below.
``M'' is the main label, and ``S'' is the sub-label.

\begin{table}[!tb]
    \caption{Search range for hyperparameters and their values.}
    \label{tab:hyper_params}
    \centering
    \begin{tabular}{r|l}
         \hline
         \hline
         Hyperparameters & Values \\
         \hline
         Label smoothing& none or $\{0.05,\underline{0.1},0.2,0.3\}$\\
         Initial LR& $\{0.005,0.01,\ldots,\underline{0.025},\ldots,0.05\}$\\
         Momentum & \underline{$0.9$}\\
         Weight decay & \{1e-5, 5e-5, \underline{1e-4}, 5e-4\}\\
         \hline
         \hline
    \end{tabular}
    \begin{tablenotes}
     \footnotesize
     \item[*]\underline{Underlines} indicate adopted values.
    \end{tablenotes}
\end{table}

\subsection{Experimental settings}
\label{sec:ex_performance}
We implemented our experimental model in Python and used the PyTorch 1.11.0+cu102 framework on one machine with a single NVIDIA V100 GPU.
Each experiment typically took several hours to complete.

We confirmed the performance of our experimental model using the dataset described in the previous section.
To have the same number of samples for each label, we set ($S$) as an upper limit on the number of random samples from each label used in the evaluation experiment.

Our experiments had hyperparameters to be tuned, so we first considered the values in Table~\ref{tab:hyper_params} when searching for appropriate values of them.
Cosine learning rate decay~\cite{loshchilov2016sgdr} was adopted with an initial learning rate of 0.025.
Label smoothing~\cite{muller2019does} was also adopted with a factor of 0.1.
Data augmentation was not employed.
The other experimental conditions were identical to those employed in o-glassesX, allowing for comparison with existing methods; accuracies were obtained using 4-fold cross-validation, and the other parameter configurations were as follows.
\begin{itemize}
 \item input length ($L$) $ = 235$ bytes
 \item $S$ (Num. of random samples from each label) $= 20000$
 \item mini-batch size $ = 64$
 \item epochs $ = 50$
\end{itemize}

\begin{table*}[!tb]
\caption{Comparison of learning methods for the centrifuge mechanism.}
\label{tab:ex1_3}
    \centering
    \begin{tabular}[t]{cc|ccccc|c}
         \hline
         \hline
         \multicolumn{2}{r|}{Learning method}& UTL & DTL & UFT & DFT & 2LF & baseline\\
         \hline
         \multicolumn{7}{l}{Ex.1: Main(51:compiler with opt. lv.), Sub(9:CPU architecture)}\\
         \cline{3-8}
         &Main Acc.& 97.08$_{\pm0.07}$ & \underline{96.92}$_{\pm0.06}$ & 97.12$_{\pm0.04}$ & \textbf{97.22}$_{\pm0.07}$ & 97.01$_{\pm0.03}$& 97.13$_{\pm0.07}$\\
         &Sub Acc.& 99.66$_{\pm0.02}$ & 99.59$_{\pm0.02}$ & 96.33$_{\pm0.09}$ & \underline{54.79}$_{\pm14.38}$ & \textbf{99.69}$_{\pm0.03}$& 10.14$_{\pm1.11}$\\
         &\raisebox{0.75cm}{\shortstack{$\mathbf{x'}$\\(CPU Arch.)}} & \includegraphics[width=1.5cm]{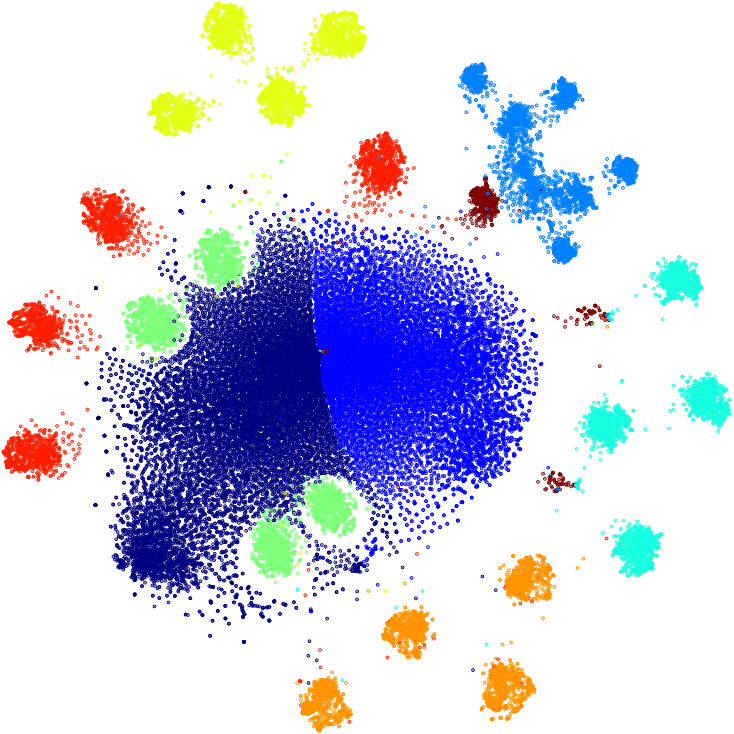}& \includegraphics[width=1.5cm]{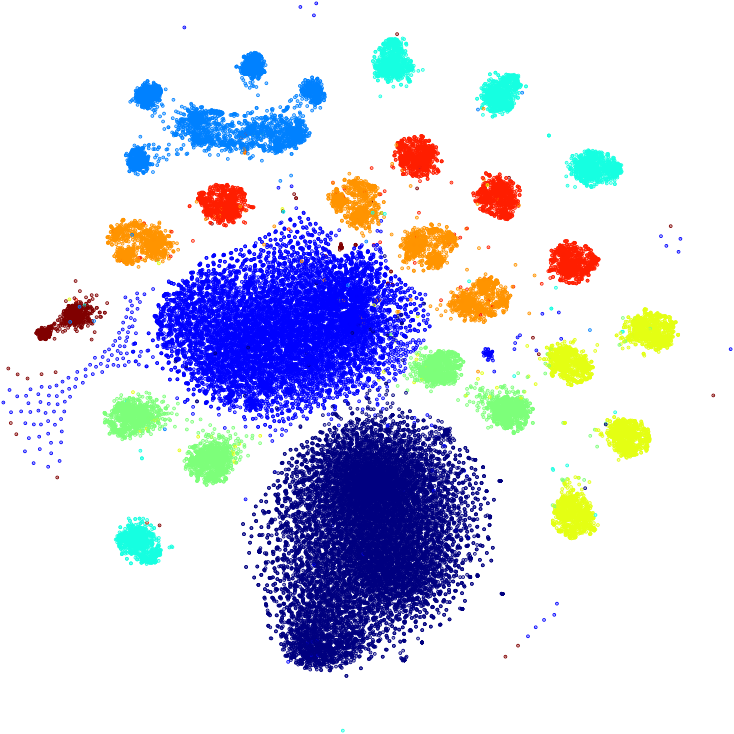}&\includegraphics[width=1.5cm]{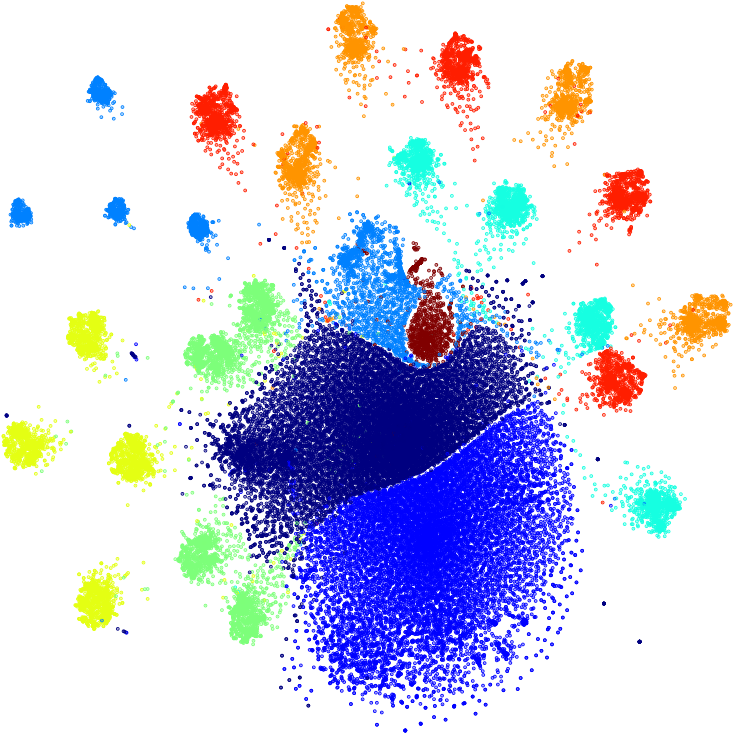} &\includegraphics[width=1.5cm]{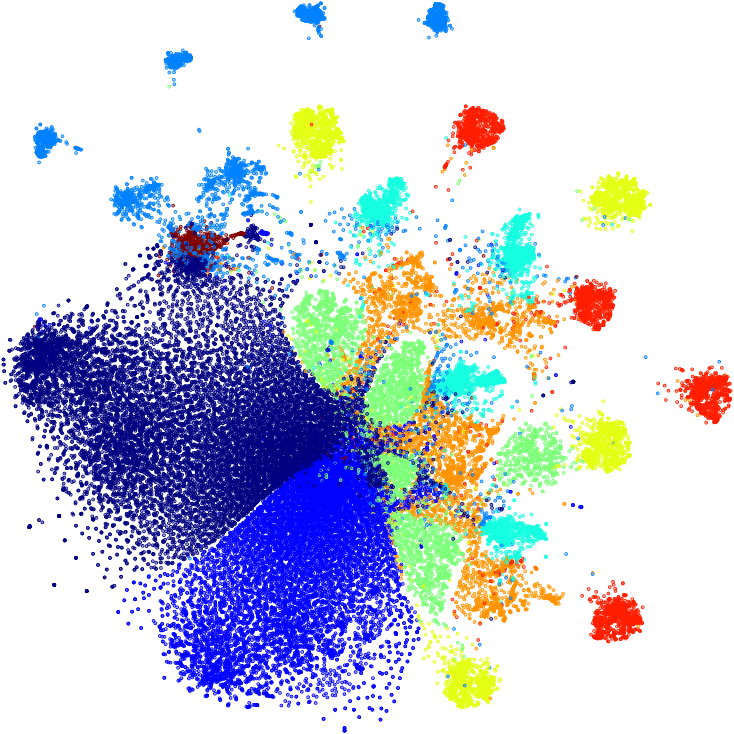} &\includegraphics[width=1.5cm]{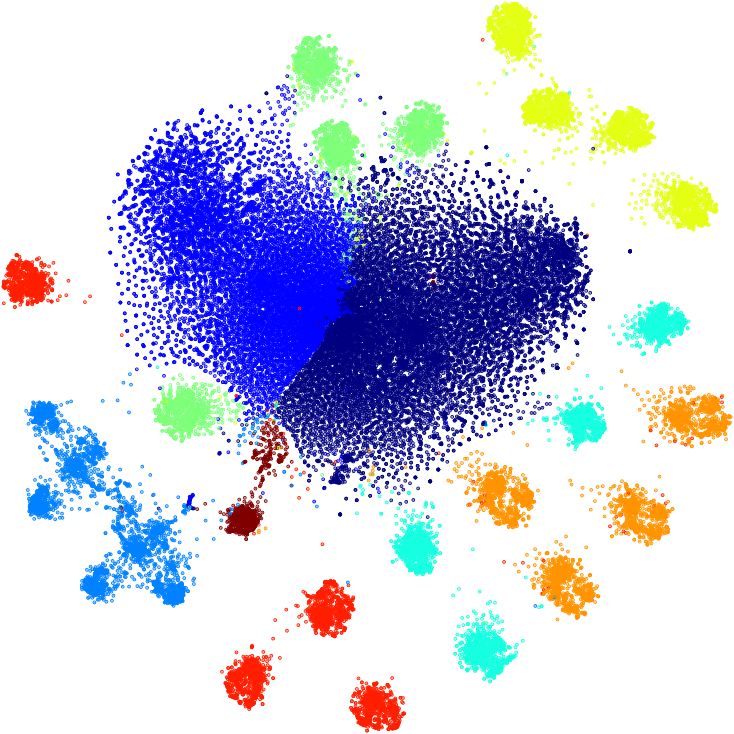} &\includegraphics[width=1.5cm]{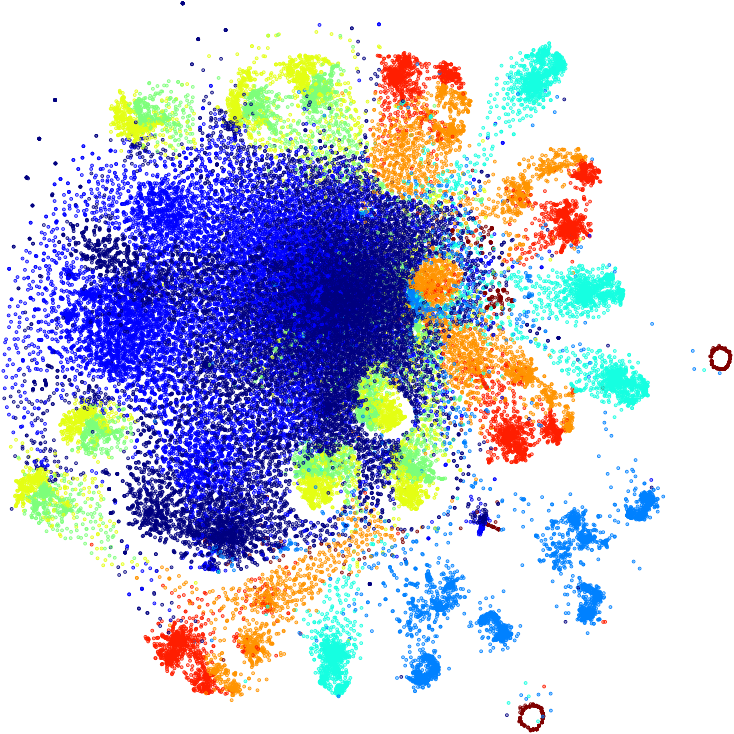}\\
         \hline
         \multicolumn{7}{l}{Ex.2: Main(9:CPU architecture), Sub(51:compiler with opt. lv.)}\\
         \cline{3-8}
         &Main Acc.& 99.74$_{\pm0.02}$ & \underline{99.62}$_{\pm0.01}$ & \textbf{99.79}$_{\pm0.01}$ & 99.66$_{\pm0.01}$ & 99.78$_{\pm0.01}$ & 99.68$_{\pm0.02}$\\
         &Sub Acc.& \textbf{97.09}$_{\pm0.03}$ & 18.29$_{\pm1.25}$ & 20.66$_{\pm0.92}$ & \underline{1.64}$_{\pm0.18}$ & 97.08$_{\pm0.03}$ & 1.64$_{\pm0.95}$\\
         &Arch. Acc.& 99.72$_{\pm0.02}$ & 99.58$_{\pm0.01}$ & 99.48$_{\pm0.15}$ & \underline{12.23}$_{\pm6.87}$ & \textbf{99.73}$_{\pm0.01}$ & 15.26$_{\pm11.61}$\\
         &\raisebox{0.75cm}{\shortstack{$\mathbf{x'}$\\(CPU Arch.)}} & \includegraphics[width=1.5cm]{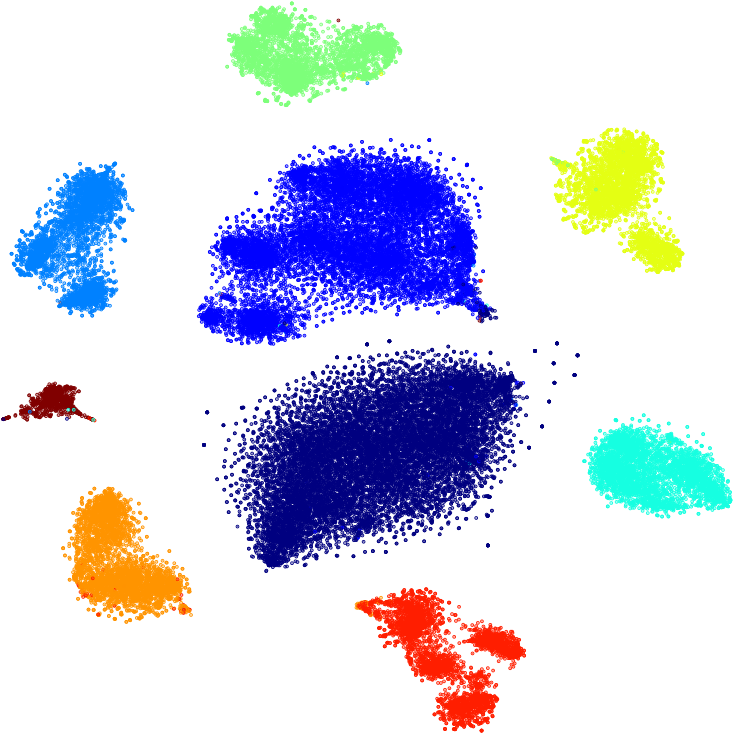}&\includegraphics[width=1.5cm]{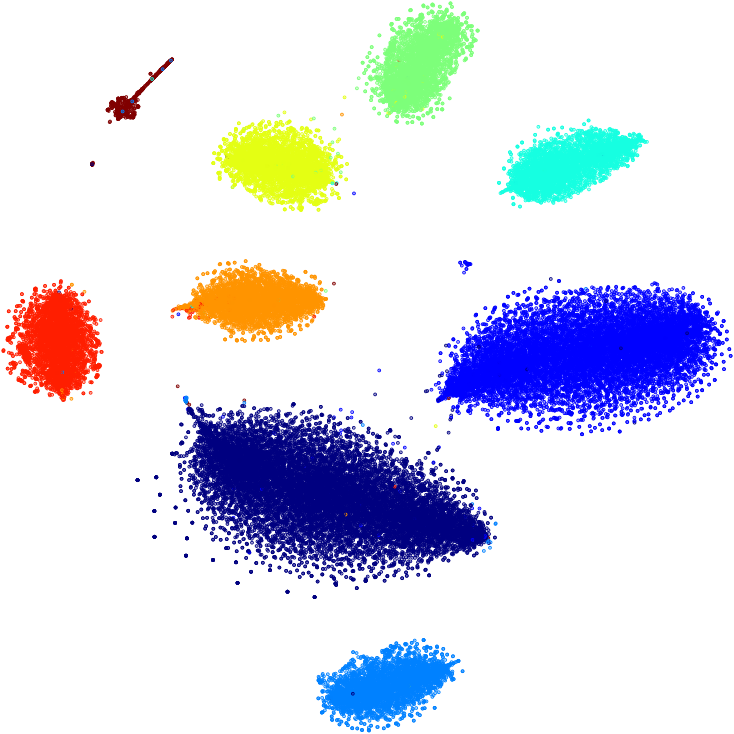}&\includegraphics[width=1.5cm]{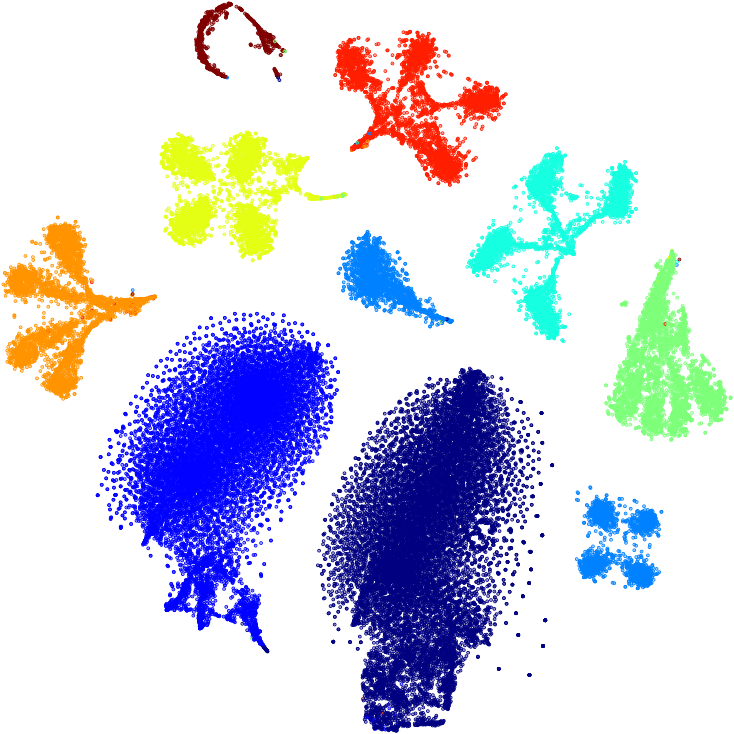} &\includegraphics[width=1.5cm]{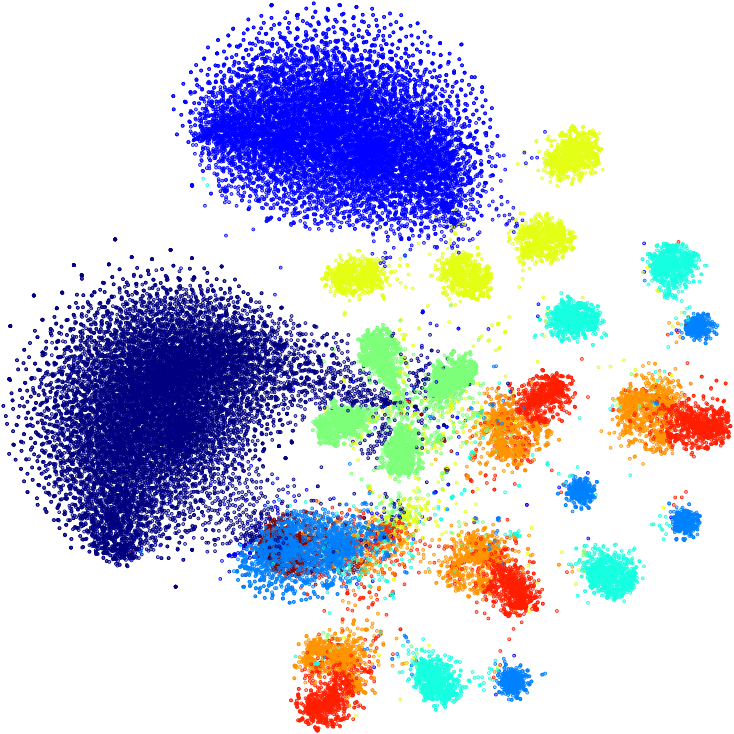} &\includegraphics[width=1.5cm]{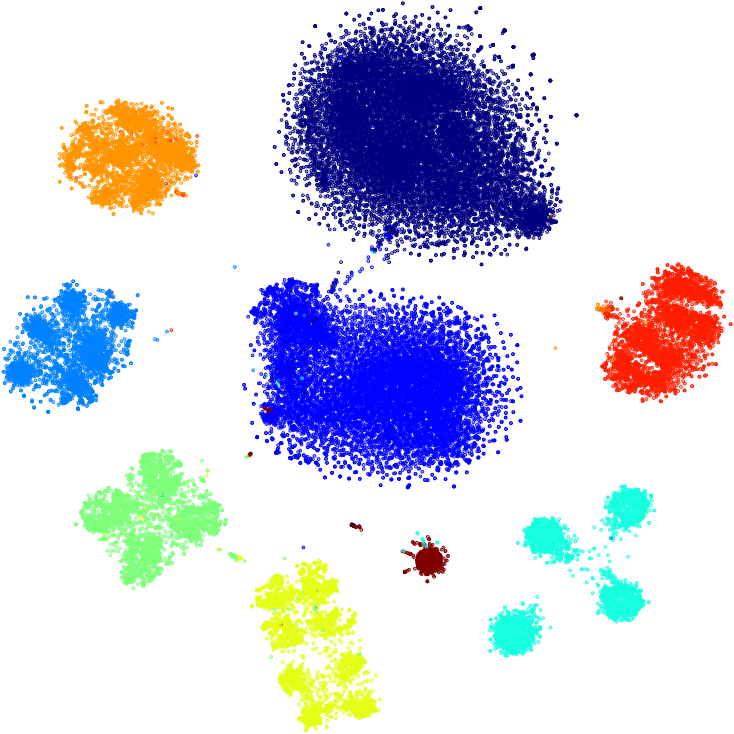} &\includegraphics[width=1.5cm]{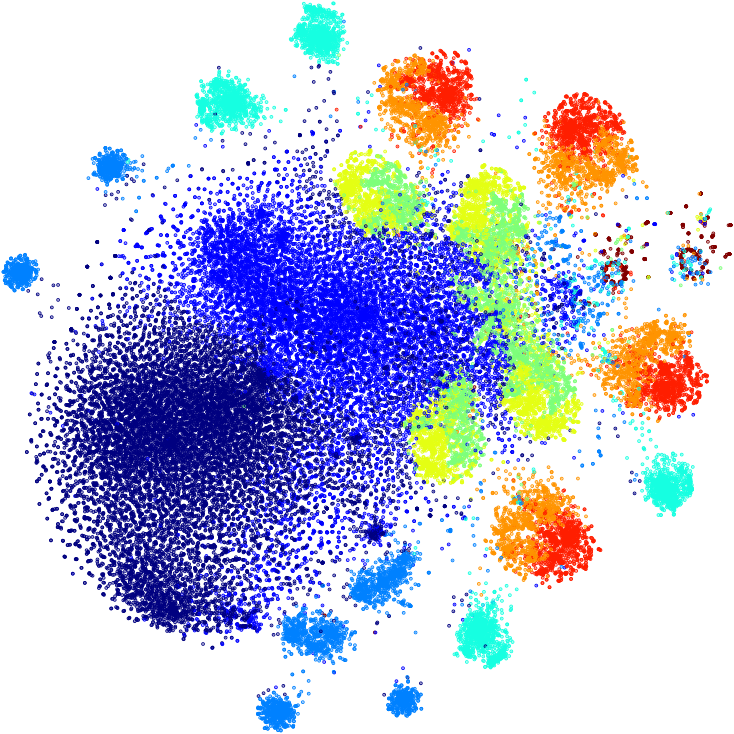}\\
         \hline
         \multicolumn{7}{l}{Ex.3: Main(26:compiler without opt. lv.), Sub(3:opt. lv.)}\\
         \cline{3-8}
         &Main Acc.& 96.70$_{\pm0.12}$ & 96.72$_{\pm0.03}$ & \underline{96.35}$_{\pm0.11}$ & \textbf{97.00}$_{\pm0.04}$ & 96.72$_{\pm0.11}$ & 96.56$_{\pm0.10}$\\
         &Sub Acc.& 99.66$_{\pm0.03}$ & 98.04$_{\pm0.23}$ & 40.91$_{\pm4.14}$ & \underline{20.87}$_{\pm1.64}$ & \textbf{99.69}$_{\pm0.01}$ & 43.26$_{\pm5.38}$\\
         &\raisebox{0.75cm}{\shortstack{$\mathbf{x'}$\\(Opt. lv.)}} & \includegraphics[width=1.5cm]{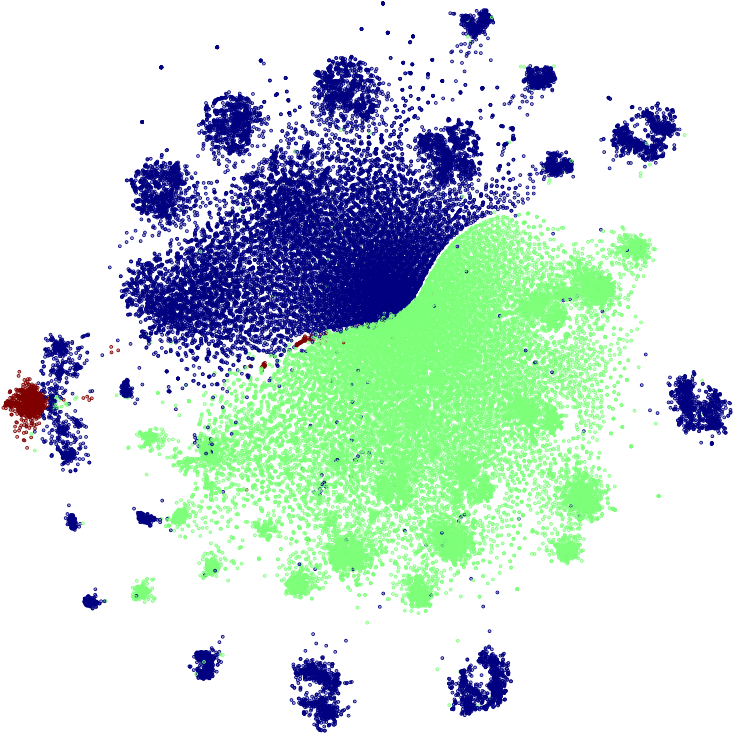}& \includegraphics[width=1.5cm]{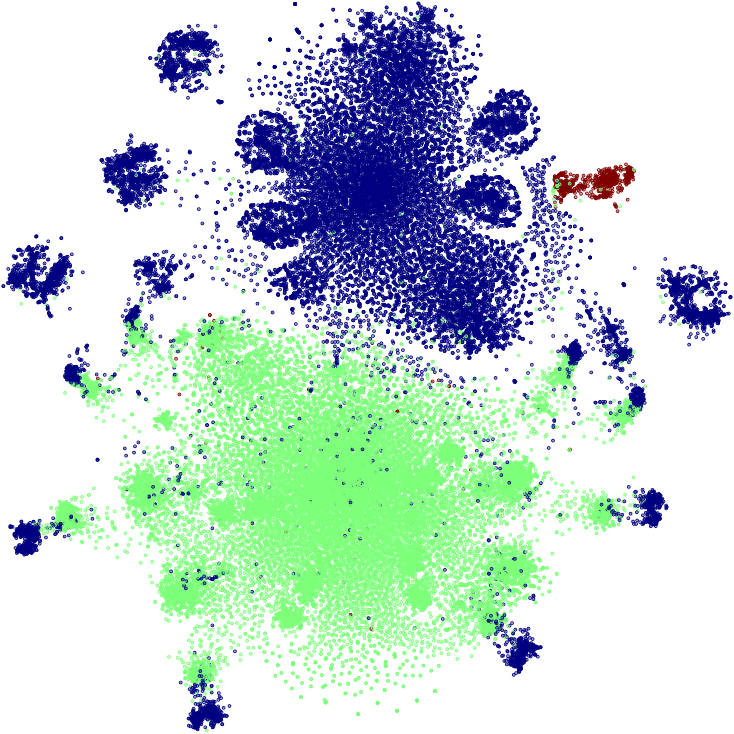}&\includegraphics[width=1.5cm]{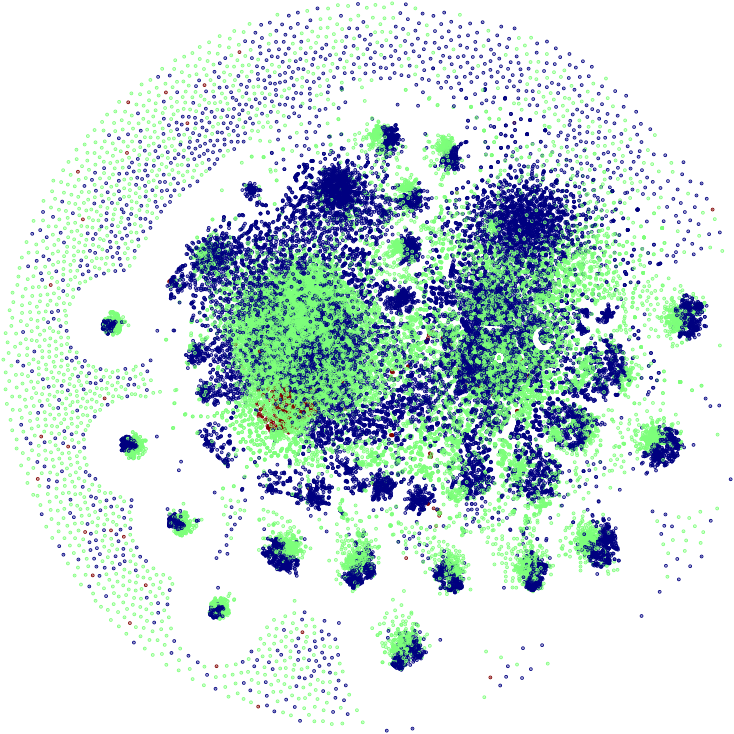} &\includegraphics[width=1.5cm]{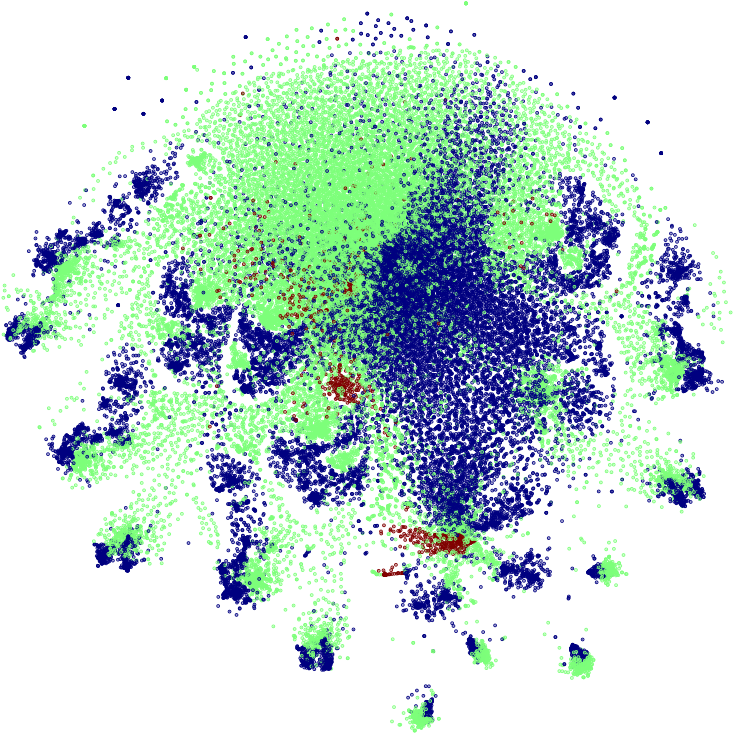} &\includegraphics[width=1.5cm]{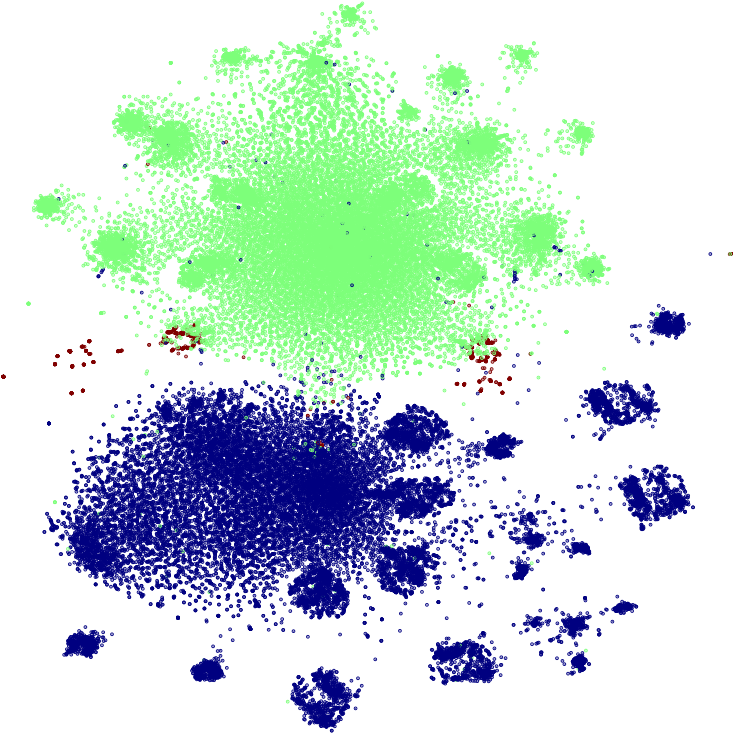} &\includegraphics[width=1.5cm]{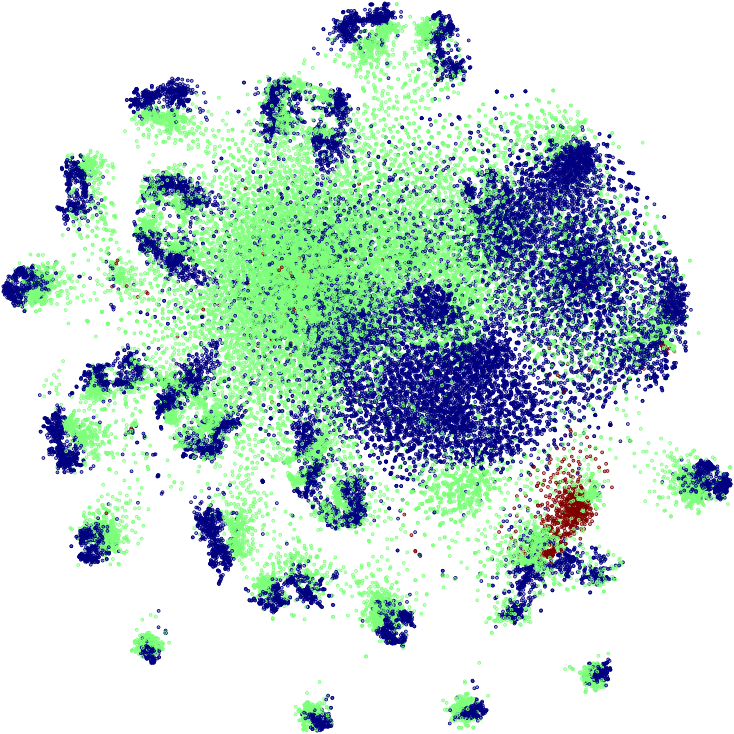}\\
         \hline
         \multicolumn{2}{r|}{Pretrain}& Sub w/ $\mathcal{L}_\mathrm{S}$& Main w/ $\mathcal{L}_\mathrm{M}$& Sub w/ $\mathcal{L}_\mathrm{S}$ & Main w/ $\mathcal{L}_\mathrm{M}$ & none & none \\
         \multicolumn{2}{r|}{Train}& Main w/ $\mathcal{L}_\mathrm{M}$& Sub w/ $\mathcal{L}_\mathrm{M}$& Both w/ $\mathcal{L}_\mathrm{M}$ & Both w/ $\mathcal{L}_\mathrm{M}$ & Both w/ $\mathcal{L}_\mathrm{M}$\&$\mathcal{L}_\mathrm{S}$ & Both w/ $\mathcal{L}_\mathrm{M}$\\
         
         \multicolumn{2}{r|}{Sub-nets contribution}& unclear& \textbf{clear}& unclear & unclear & unclear & unclear \\
         \hline
         \hline
    \end{tabular}
    \begin{tablenotes}
     \footnotesize
     \item[*]The best results under the same conditions are indicated in \textbf{bold}, and the worst results are denoted by \underline{underlined} text. $\mathbf{x'}$ is the output of the main-net first layer visualized by tSNE., and color-coding rules are described in parentheses. ``Arch. Acc.'' is the accuracy of the sub-net when the corrects and the predictions are grouped by CPU architecture. ``Sub-nets contribution'' indicates whether sub-nets contributed to the main class classification.
    \end{tablenotes}
\end{table*}

\subsection{Ex.~1 to Ex.~3: Comparison between the five learning methods}
In this section, we check the classification performance of each learning method using three different main-labels and sub-labels for our dataset.
Each labeling rule is as shown in ``Label'' in Table 1, specifically the following three labeling rules.

\begin{itemize}
    \item Ex.~1: We set main labels for 51 classes of compiler identification, including optimization level identification, and sub-labels for nine classes of CPU architecture identification. We then check the classification performance of our experimental model when the sub-prediction is easier than the main prediction.

    \item Ex.~2: We swapped the main-labels and sub-labels from Ex.~1. We then check the classification performance of our experimental model with nine main-labels for the CPU architecture and 51 sub-labels for compiler identification with optimization level identification included.

    \item Ex.~3: We check the classification performance of our experimental model when the main-label and sub-label are independent. We chose 26 labels for compiler identification that did not include optimization level estimation as main-labels.

\end{itemize}

Table~\ref{tab:ex1_3} shows a comparison of the learning methods for the centrifuge mechanism and baseline, which simply trains both the main-net and sub-net with $\mathcal{L}_\mathrm{M}$.

The $\mathbf{x'}$ in Table~\ref{tab:ex1_3} shows the main-net first linear layer output visualized by tSNE~\cite{maaten2008visualizing}, color-coded by CPU architectures or optimization levels.
$\mathbf{x'}$ is indicated by Equation~\ref{eq:x_dash}, and we expect $\mathbf{x'}$ to be the input $\mathbf{x}$ transitioned by each sub-label, as shown in Figure~\ref{fig:x_dash}.

Simply comparing the accuracy of the main-label classification, we can see no significant performance difference among any of the learning methods.
On the other hand, when we focus on the accuracy of sub-label classification, we find that fine-tuning does not perform well in sub-label classification due to catastrophic forgetting~\cite{mccloskey1989catastrophic}.

Focusing on DTL in Ex.~1 and Ex.~3, we can see that the sub-prediction achieves a remarkable degree of accuracy, even though only the loss function for the main-label is used.

In Ex.~2, the sub-predictions are not well classified except for UTL and 2LF.
One possible cause is that one main-label has a relationship that encompasses multiple sub-labels.
If the main label classification is known to belong to one of those multiple sub-labels, then it can be classified without identifying the difference between them.
Therefore, it is likely that error propagation from the main-net to the sub-net did not work to improve the accuracy of sub-label classification.
From the main classification standpoint, the accuracy of the sub-label classification does not need to be perfect, it only needs to be correct at the CPU architecture level for the main-label.
DTL's $\mathbf{x'}$ of Ex.~2 in the table clearly shows that $\mathbf{x'}$ is limited to the essence for the main prediction, that is, CPU architecture identification.
Therefore, the sub-net accuracy of DTL may measure the importance of sub-label classification for main-label classification.

In summary, focusing on DTL in particular, we found the following characteristics compared with other learning methods.
\begin{itemize}
    \item DTL can incorporate highly accurate sub-predictions into the model without having to adjust hyperparameters or design loss functions for sub-predictions, such as 2LF's $\beta$.

    \item The accuracy of DTL sub-predictions depends on the essence of the main forecast, and it might be used to measure their contribution to the main prediction.

\end{itemize}

\subsubsection{Ex.~4: Classification performance in the case of a model containing two sub-nets}
The results of Ex.~1 to Ex.~3 were those for which the model contained a single sub-net.
This section will examine the case where the number of sub-nets is increased to two.
As with the other experiments, o-glassesX was used for the main-net and the two sub-nets.
The main-labels are for compiler identification with optimization level identification, and there are 51 of them.
The first sub-labels are three labels for optimization level identification.
The second sub-labels are nine labels for CPU architecture identification.
The output of the two sub-nets and the input $\mathbf{x}$ are combined and put into the main-net.

Based on the results of Ex.~1 to Ex.~3, we employed DTL as the learning method since there is no need to design a loss function for sub-predictions.

Table~\ref{tab:ex_4} shows the results of Ex.~4.
\begin{table}[!tb]
    \caption{The results of Ex.~4.}
    \label{tab:ex_4}
    \centering
    \begin{tabular}{c|ccc}
         \hline
         \hline
          & Main & Sub$_1$& Sub$_2$ \\
         \hline
         DTL& 96.72$_{\pm0.05}$ & 99.62$_{\pm0.03}$ & 99.56$_{\pm0.02}$\\ 
         \hline
         \hline
    \end{tabular}
\end{table}
The experimental results show that even when the number of sub-nets is increased to two, sub-predictions can be made with high accuracy in DTL.
Thus, it may be easier to develop a model with more complex coordination of sub-nets.

\subsection{Ex.~5: Performance comparison with existing methods (only x86/x86-64)}
In this section, we compare our proposed and existing methods' classification accuracy.
We have to compare our proposed method with existing CPR methods that support multiple CPU architectures.
To our knowledge, however, our method is the only one that supports multiple CPU architectures.
In most studies, the classification targets are programs with an x86/x86-64 architecture.
Therefore, we conducted a comparison experiment using only the x86/x86-64 architecture programs among our dataset (only those marked with a checkmark in Table~\ref{tab:dataset}.)
We set three architectures (x86/x86-64/Others) as sub-labels in our method
and selected DFT as the learning method.
Table~\ref{tab:ex_mono} shows a comparison between our method and other existing methods.

\begin{table}[!tb]
\caption{Performance comparison with related work (on x86/x86-64).} 
\label{tab:ex_mono}
\begin{center}
\scalebox{0.65}{
\begin{sc}
\begin{tabular}{r|ccccc}
\hline
\hline
Method & Acc. & ML model & No DE & Features & \#Labels\\
\hline
Ours (DFT) & 98.62 & Attention & \checkmark & 235~bytes & 19\\
o-glassesX & \textbf{98.84} & Attention & & 64 Instructions$^{[1]}$ & 19\\
o-glasses & 94.21 & CNN & & 64 Instructions$^{[1]}$ & 19\\
CNN & 36.33 & CNN & \checkmark & 235~bytes & 19\\
\hline
Rosenblum's & 92.4 & CRF &  & Instruction Seq. & 3\\
ORIGIN (SVM) & 60.4 & SVM &  & 1 Function & 18\\
ORIGIN (CRF) & 91.8 & CRF &  & Function Seq. & 18\\
BinComp & 80.1 & (k-means) &  & 1 File & 6\\
\hline
\hline
\end{tabular}
\end{sc}
}
    \begin{tablenotes}
     \footnotesize
     \item[*]The best result is indicated in \textbf{bold}. [1]: The average size of 64 instructions is 236 bytes.
    \end{tablenotes}
\end{center}
\end{table}

In Table~\ref{tab:ex_mono}, ``CNN'' has the same model as o-glasses' but does not perform pre-processing for instruction segmentation.
For existing methods that cannot be tested on our dataset due to different feature vectors, the classification accuracies are taken from their papers (\cite{rosenblum2010extracting}, \cite{rosenblum2011recovering} and \cite{rahimian2015bincomp}).

The only difference between CNN and o-glasses is whether or not DE is performed.
However, there are significant differences in performance between the two methods.
This difference in performance indicates that DE plays a very important role in CPR.
On the other hand, even though our method does not perform DE, there is little performance difference from o-glassesX, which does so.
Therefore, we believe that the centrifuge mechanism can replace DE.
By not relying on DE, the model's range of applicability may be expanded, because it can be trained on a mixed data set such as ours.

\section{Conclusion}
We proposed a centrifuge mechanism in which the upstream sub-net transitions the input to a space corresponding to sub-labels without manual DE. 
Centrifuge allows the downstream main-net to focus more on difficult main-label classifications.
DTL, one of the learning methods for the centrifuge mechanism, pre-trains the main-net using the sub-label's ground truth instead of the sub-net's output.
DTL was found to be able to give the sub-net the role of sub-label prediction without using a loss function for sub-label classification.
Additionally, we found that sub-predictions tend to be highly accurate when the sub-label classification contributes to the essence of the main prediction.
We applied the centrifuge mechanism to CPR.
Our experiments confirmed that our experimental model without DE could classify bit-streams with an accuracy of 98.62, the same high performance as existing CPR methods with DE for x86/x86-64 CPU architecture.
Also, we applied our model to compiler identification for multiple CPU architectures and confirmed that our model could classify 51 classes with an accuracy of 97.22.
To our knowledge, our compiler identification model is the first that can simultaneously support multiple CPU architectures, and it has achieved top-class classification performance for compiler identification.



\bibliography{ref}

\begin{thebibliography}{18}
\providecommand{\natexlab}[1]{#1}

\bibitem[{Cortes and Vapnik(1995)}]{cortes1995support}
Cortes, C.; and Vapnik, V. 1995.
\newblock Support-vector networks.
\newblock \emph{Machine Learning}, 20(3): 273--297.

\bibitem[{Girshick et~al.(2014)Girshick, Donahue, Darrell, and
  Malik}]{girshick2014rich}
Girshick, R.; Donahue, J.; Darrell, T.; and Malik, J. 2014.
\newblock Rich feature hierarchies for accurate object detection and semantic
  segmentation.
\newblock In \emph{Proceedings of the IEEE Conference on Computer Vision and
  Pattern Recognition}, 580--587. New York City, NY: IEEE.

\bibitem[{Heo et~al.(2019)Heo, So, Yang, Yoon, and Yu}]{Heo2019}
Heo, H.-S.; So, B.-M.; Yang, I.-H.; Yoon, S.-H.; and Yu, H.-J. 2019.
\newblock Automated recovery of damaged audio files using deep neural networks.
\newblock \emph{Digital Investigation}, 30: 117--126.

\bibitem[{Loshchilov and Hutter(2016)}]{loshchilov2016sgdr}
Loshchilov, I.; and Hutter, F. 2016.
\newblock Sgdr: Stochastic gradient descent with warm restarts.
\newblock \emph{arXiv:1608.03983}.

\bibitem[{McCallum(2002)}]{mccallum2002efficiently}
McCallum, A. 2002.
\newblock Efficiently inducing features of conditional random fields.
\newblock In \emph{Proceedings of the Nineteenth Conference on Uncertainty in
  Artificial Intelligence}, 403--410. Vurlington, MA: Morgan Kaufmann
  Publishers Inc.

\bibitem[{McCloskey and Cohen(1989)}]{mccloskey1989catastrophic}
McCloskey, M.; and Cohen, N.~J. 1989.
\newblock Catastrophic interference in connectionist networks: The sequential
  learning problem.
\newblock \emph{Psychology of Learning and Motivation}, 24: 109--165.

\bibitem[{M{\"u}ller, Kornblith, and Hinton(2019)}]{muller2019does}
M{\"u}ller, R.; Kornblith, S.; and Hinton, G.~E. 2019.
\newblock When does label smoothing help?
\newblock \emph{Advances in Neural Information Processing Systems}, 32:
  4696--4705.

\bibitem[{Nataraj et~al.(2011)Nataraj, Karthikeyan, Jacob, and
  Manjunath}]{nataraj2011malware}
Nataraj, L.; Karthikeyan, S.; Jacob, G.; and Manjunath, B.~S. 2011.
\newblock Malware images: visualization and automatic classification.
\newblock In \emph{Proceedings of the 8th International Symposium on
  Visualization for Cyber Security}, 1--7. New York City, NY: Association for
  Computing Machinery.

\bibitem[{{Otsubo} et~al.(2020){Otsubo}, {Otsuka}, {Mimura}, and
  {Sakaki}}]{otsubo2018glasses}
{Otsubo}, Y.; {Otsuka}, A.; {Mimura}, M.; and {Sakaki}, T. 2020.
\newblock o-glasses: Visualizing x86 code from binary using a 1D-CNN.
\newblock \emph{IEEE Access}, 8: 31753--31763.

\bibitem[{Otsubo et~al.(2020)Otsubo, Otsuka, Mimura, Sakaki, and
  Ukegawa}]{otsuboglassesx}
Otsubo, Y.; Otsuka, A.; Mimura, M.; Sakaki, T.; and Ukegawa, H. 2020.
\newblock o-glassesX: Compiler provenance recovery with attention mechanism
  from a short code fragment.
\newblock In \emph{Workshop on Binary Analysis Research}, 1--12. Reston, VA:
  Internet Society.

\bibitem[{Padmanabhuni and Tan(2015)}]{padmanabhuni2015buffer}
Padmanabhuni, B.~M.; and Tan, H. B.~K. 2015.
\newblock Buffer overflow vulnerability prediction from x86 executables using
  static analysis and machine learning.
\newblock In \emph{2015 IEEE 39th Annual Computer Software and Applications
  Conference}, 450--459. New York City, NY: IEEE.

\bibitem[{Pan and Yang(2009)}]{pan2009survey}
Pan, S.~J.; and Yang, Q. 2009.
\newblock A survey on transfer learning.
\newblock \emph{IEEE Transactions on knowledge and data engineering}, 22(10):
  1345--1359.

\bibitem[{Rahimian et~al.(2015)Rahimian, Shirani, Alrabaee, Wang, and
  Debbabi}]{rahimian2015bincomp}
Rahimian, A.; Shirani, P.; Alrabaee, S.; Wang, L.; and Debbabi, M. 2015.
\newblock BinComp: A stratified approach to compiler provenance attribution.
\newblock \emph{Digital Investigation}, 14 (Suppl 1): S146--S155.

\bibitem[{Rosenblum, Miller, and Zhu(2011)}]{rosenblum2011recovering}
Rosenblum, N.; Miller, B.~P.; and Zhu, X. 2011.
\newblock Recovering the toolchain provenance of binary code.
\newblock In \emph{Proceedings of the 2011 International Symposium on Software
  Testing and Analysis}, 100--110. New York City, NY: Association for Computing
  Machinery.

\bibitem[{Rosenblum et~al.(2007)Rosenblum, Zhu, Miller, and
  Hunt}]{rosenblum2007machine}
Rosenblum, N.; Zhu, X.; Miller, B.; and Hunt, K. 2007.
\newblock Machine learning-assisted binary code analysis.
\newblock In \emph{NIPS Workshop on Machine Learning in Adversarial
  Environments for Computer Security}, 1--3. Whistler, Canada.

\bibitem[{Rosenblum, Miller, and Zhu(2010)}]{rosenblum2010extracting}
Rosenblum, N.~E.; Miller, B.~P.; and Zhu, X. 2010.
\newblock Extracting compiler provenance from program binaries.
\newblock In \emph{Proceedings of the 9th ACM SIGPLAN-SIGSOFT Workshop on
  Program Analysis for Software Tools and Engineering}, 21--28. New York City,
  NY: Association for Computing Machinery.

\bibitem[{Rosenblum, Zhu, and Miller(2011)}]{Rosenblum2011WhoWT}
Rosenblum, N.~E.; Zhu, X.; and Miller, B. 2011.
\newblock Who wrote this code? Identifying the authors of program binaries.
\newblock In \emph{ESORICS 2011}, 172--189. Berlin, Germany: Springer.

\bibitem[{Van~der Maaten and Hinton(2008)}]{maaten2008visualizing}
Van~der Maaten, L.; and Hinton, G. 2008.
\newblock Visualizing data using t-SNE.
\newblock \emph{Journal of Machine Learning Research}, 9: 2579--2605.

\end{thebibliography}

\end{document}